\documentclass[conference]{IEEEtran}
\IEEEoverridecommandlockouts
\usepackage{cite}
\usepackage{amsmath,amssymb,amsfonts}
\usepackage{algorithmic}
\usepackage{graphicx}
\usepackage{textcomp}
\usepackage{xcolor}

\usepackage{tabularx}
\usepackage{booktabs}
\usepackage{hyperref}

\usepackage[pscoord]{eso-pic}

\def\BibTeX{{\rm B\kern-.05em{\sc i\kern-.025em b}\kern-.08em
    T\kern-.1667em\lower.7ex\hbox{E}\kern-.125emX}}
\begin{document}

\title{A Spatio-Temporal Attentive Network for Video-Based Crowd Counting\\
\thanks{This work was partially funded by: Extension (ESA, n. 4000132621/20/NL/AF), AI4Media - A European Excellence Centre for Media, Society and Democracy (EC, H2020 n. 951911), “Intelligenza Artificiale per il Monitoraggio Visuale dei Siti Culturali" (AI4CHSites) CNR4C program, CUP B15J19001040004, and by the Italian Ministry of University and Research MUR-FOE 2020 TIRS project.}
}

\author{\IEEEauthorblockN{1\textsuperscript{st} Marco Avvenuti}
\IEEEauthorblockA{\textit{Dept. of Information Engineering} \\
\textit{University of Pisa}\\
Pisa, Italy \\
marco.avvenuti@unipi.it}
\and
\IEEEauthorblockN{2\textsuperscript{nd} Marco Bongiovanni}
\IEEEauthorblockA{\textit{Dept. of Information Engineering} \\
\textit{University of Pisa}\\
Pisa, Italy \\
m.bongiovanni2@studenti.unipi.it}
\and
\IEEEauthorblockN{3\textsuperscript{rd} Luca Ciampi}
\IEEEauthorblockA{\textit{Inst. of Information Science and Tech.} \\
\textit{National Research Council (ISTI-CNR)}\\
Pisa, Italy \\
luca.ciampi@isti.cnr.it}
\and
\IEEEauthorblockN{4\textsuperscript{th} Fabrizio Falchi}
\IEEEauthorblockA{\textit{Inst. of Information Science and Tech.} \\
\textit{National Research Council (ISTI-CNR)}\\
Pisa, Italy \\
fabrizio.falchi@isti.cnr.it}
\and
\IEEEauthorblockN{5\textsuperscript{th} Claudio Gennaro}
\IEEEauthorblockA{\textit{Inst. of Information Science and Tech.} \\
\textit{National Research Council (ISTI-CNR)}\\
Pisa, Italy \\
claudio.gennaro@isti.cnr.it}
\and
\IEEEauthorblockN{6\textsuperscript{th} Nicola Messina}
\IEEEauthorblockA{\textit{Inst. of Information Science and Tech.} \\
\textit{National Research Council (ISTI-CNR)}\\
Pisa, Italy \\
nicola.messina@isti.cnr.it}
}

\maketitle

\begin{abstract}
Automatic people counting from images has recently drawn attention for urban monitoring in modern Smart Cities due to the ubiquity of surveillance camera networks. Current computer vision techniques rely on deep learning-based algorithms that estimate pedestrian densities in still, individual images. Only a bunch of works take advantage of temporal consistency in video sequences. In this work, we propose a spatio-temporal attentive neural network to estimate the number of pedestrians from surveillance videos. By taking advantage of the temporal correlation between consecutive frames, we lowered state-of-the-art count error by $5\%$ and localization error by $7.5\%$ on the widely-used FDST benchmark.
\end{abstract}


\section{Introduction}
\label{sec:introduction}
Computer Vision obtained a tremendous boost in the last few years thanks to the astonishing advances in Machine Learning. In particular, Deep Learning allowed the research community to define new state-of-the-arts in many Computer Vision tasks, such as object detection \cite{DBLP:conf/iscc/TraquairKKK19}, or image retrieval \cite{messina_1}, to name a few.
With the increasing interest in Smart Cities and the grown availability of surveillance cameras that have become pervasive, there is a unanimous effort to employ these novel technologies for urban monitoring and surveillance. Like no other sensing mechanism, networks of city cameras can observe and simultaneously provide visual data to AI systems to extract relevant information from this deluge of data. In this context, many smart applications, ranging from lot occupancy detection \cite{amato_lot_occupancy} to pedestrian detection \cite{viped_1, viped_2, Di_Benedetto_2022} and re-identification \cite{DBLP:conf/iscc/MaBZLHL21}, have been proposed and are nowadays widely employed worldwide. 
In this work, we treat the \textit{counting} task, which consists of providing
the number of instances of a specific class present in the scene --- for example, the number of vehicles that are transiting a particular road. Specifically, we focus on crowd counting to automatically estimate the number of people present in images gathered from city surveillance cameras. This application is crucial in many scenarios, like monitoring and eventually limiting people aggregations during the recent COVID-19 pandemic.

Recent literature extensively faced the crowd counting task by estimating and integrating density maps of still, individual images. Nevertheless, relatively few works take advantage of the temporal consistency of \textit{video} streams. However, using temporal constraints across consecutive frames could be an essential key point for enhancing the counting performance. 


Driven by these concerns, we propose an enhanced extension to the video-counting framework introduced by Liu et al. \cite{liu2020estimating, 9508171}. These works presented an interesting semi-supervised learning method that infers people density maps starting from the estimation of people flows among adjacent keyframes. At inference time, the flow is integrated to obtain the actual people count. Although the learning framework is solid, their proposed network comprises a simple fully-convolutional encoder-decoder pipeline, which estimates the flow from a pair of consecutive images. In this work, inspired by the recent attentive mechanisms proposed to process visual data, such as Vision Transformer \cite{vit}, we enhance the architecture in \cite{liu2020estimating, 9508171} with self-attentive connections. Specifically, we introduce an attentive-based temporal fusion layer to improve the flow predictor, taking advantage of the temporal correlation between consecutive frames. Through an experimental evaluation, we show that our proposed method can boost the performance compared with the original framework on the widely-used FDST dataset \cite{DBLP:conf/icmcs/FangZCGH19}, one of the largest and most diverse collections of temporally correlated frames, suitable for video-based counting. We assess not only the counting performance considering the counting errors occurring at inference time (i.e., the difference between the predicted and the actual person numbers) but also the ability to correctly localize the counted persons. Indeed, count errors do not take into account \textit{where} the pedestrians have been detected in the images and, consequently, counting models might achieve low values of errors while providing wrong predictions (e.g., a high number of false positives and false negatives).

To sum up, we propose the following contributions:
\begin{itemize}
    \item We propose an extension to a recent semi-supervised video-based counting framework \cite{liu2020estimating, 9508171}, employing an attentive-based temporal fusion layer that takes advantage of the temporal correlation between consecutive images and improves the people flow estimation.
    \item We demonstrate through detailed experiments that the proposed variation can reach state-of-the-art results on the FDST dataset, lowering the count error by $5\%$.
    \item We conduct a performance evaluation also considering the ability to correctly localize the counted persons, lowering the localization error by $7.5\%$ compared to the original counting framework.
\end{itemize}

\def\UrlFont{\bfseries}
The code and the trained models will be publicly available at \url{https://tinyurl.com/yb42ce38}.


\section{Related Work}
\label{sec:relatedwork}

\subsection{Image-based Counting}
Image-based counting aims at estimating the number of object instances, like people \cite{Liu_2019, multi_column}, cells \cite{ciampi_cells, Cohen_2017}, or vehicles \cite{amato2019, ciampi_multi_camera}, in \textit{still} images or video frames \cite{lempitsky2010learning}. Current solutions are formulated as supervised deep learning-based problems belonging to one of two main categories: counting by \textit{detection} and counting by \textit{regression}. Detection-based approaches, such as in \cite{amato2018wireless} and \cite{Laradji_2018}, require prior detection of the single instances of objects. On the other hand, regression-based techniques like \cite{ciampi_visapp} and \cite{can} try to establish a direct mapping between the image features and the number of objects in the scene, either directly or via the estimation of a density map (i.e., a continuous-valued function). Regression techniques show superior performance in crowded and highly-occluded scenarios \cite{lempitsky2010learning}.

\subsection{Video-based Counting}
Crowd counting approaches are mainly based on single-image inputs, even when a video sequence is available, leading to the impossibility of exploiting the temporal interdependence between consecutive frames in the sequence. Nevertheless, in the literature, there are a handful of works that rely on the estimation of density maps, but, on the other hand, try to exploit also temporal information to improve counting accuracy. For instance, \cite{xiong2017spatiotemporal} introduced one of the first video-based counting approaches, based on an LSTM-based method called ConvLSTM, i.e., a fully connected LSTM extended with convolutional layers in both the input-to-state and state-to-state connections. In \cite{fang2019locality}, the authors exploited a Locality-constrained Spatial Transformer (LST) module to model the spatial-temporal correlation between estimated neighboring density maps. 
Another remarkable work is \cite{wu2020fast}, where the authors introduced a Temporal Aware Network (TAN), which can compute the number of people present at frame $X_t$ by exploiting frames from $X_{t-k}$ to $X_{t+k}$. More recently, Liu et al. \cite{liu2020estimating, 9508171} proposed an alternative approach where density maps are not directly regressed from images but are inferred from people flows across image locations between consecutive frames. However, the proposed network in charge of estimating flows from pair of images lies in a simple fully-convolutional encoder-decoder pipeline. In this work, we extend and enhance this framework by exploiting self-attentive connections, introducing an attentive-based temporal fusion to enhance the flow predictor.

\subsection{Attentive Models}
Attention mechanisms have been largely used in the last two years and had a significant impact on tasks that involve both vision and language, such as VQA \cite{malinowski2018learning}, image captioning \cite{he2020image,cornia2020meshed} or image-text matching \cite{messina2021transformer,messina_1}. Recently, the Transformer-like attention mechanism \cite{vaswani2017attention} obtained the best results in processing images and videos. In particular, the authors in \cite{vit} introduced the Vision Transformer, demonstrating the power of the self-attentive mechanism in the image classification task. Similarly, the DETR architecture \cite{carion2020end} used the full Transformer architecture for tackling the object detection task, obtaining remarkable results with respect to state-of-the-art fully-convolutional approaches. This work is inspired by the recent advances in attentive video processing, where most methods use spatio-temporal attention to understand frame patches from multiple timesteps \cite{bertasius2021space,arnab2021vivit}.

\section{Method}
\label{sec:method}
The proposed architecture tries to reconstruct the people flows, enhancing the weak-supervised learning framework proposed in \cite{liu2020estimating, 9508171}.
In Section \ref{sec:people-flow} we briefly review their original framework; in Section \ref{sec:network}
we explain our attentive-based temporal fusion used to improve the flow predictor.

\begin{figure}[t]
  \centering
  \includegraphics[width=0.42\textwidth]{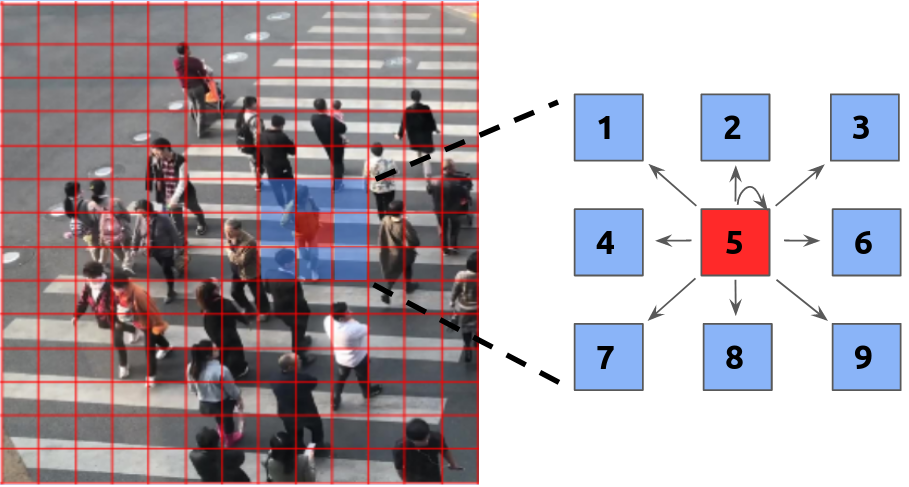}
  \caption{Visualization of a neighborhood $N(j)$ of an image patch j (shown in red). This is used to impose people conservation constraints, as formulated by Equations \ref{eq:conservation} and \ref{eq:consistency}.}
  \label{fig:flows}
\end{figure}

\subsection{The People-Flow Approach}
\label{sec:people-flow}
The approach presented in \cite{liu2020estimating, 9508171} is a video-based counting scheme that does not directly estimate people densities from images but infers them from the so-called \textit{people flow} between two consecutive frames.
People flows are vector fields that associate pedestrian movement vectors to every point in the frame space. People flows are zeros at a certain point in space if there are no moving pedestrians.

Once the people flow $\boldsymbol{f}^{t-1, t}$ between two consecutive frames $\boldsymbol{I}^{t-1}$ and $\boldsymbol{I}^{t}$ has been predicted, the j-th spatial location of the density map $d^t_{j}$ for the frame $\boldsymbol{I}^{t}$ can be reconstructed by summing all the flow contributions entering $j$ from neighboring locations of the previous frame:
\begin{align}
    d^t_{j} = \sum_{i\in N(j)} f^{t-1, t}_{i, j}
\end{align}
where the neightbouring locations $N(j)$ of the $j$-th path are shown in Figure \ref{fig:flows}.
The final people count at time $t$ can then be found by summing up all the pixel values of the obtained density map: $\sum_{j} d^t_{j}$.
However, regressing the flows is not a straightforward operation, as many video-counting datasets do not embed any explicit people flow ground-truth that can be used to supervise the network.

For this reason, the work in \cite{liu2020estimating, 9508171} proposed a weak-supervised learning approach that estimates the flows using only the ground-truth density maps $\boldsymbol{\bar{d}}^{t-1}$ and $\boldsymbol{\bar{d}}^{t}$ at consecutive timesteps $(t-1, t)$. In particular, the flows are constructed by only imposing strong people conservation constraints, i.e., people cannot appear or disappear between consecutive frames if not in the frame edges. In particular, the constraints can be expressed as follows:
\begin{align}
    \sum_{i \in N(j)}f_{i, j}^{t-1,t} &= \sum_{k \in N(j)}f_{j, k}^{t,t + 1} \label{eq:conservation}\\
    f_{i,j}^{t-1,t} &= f_{j,i}^{t,t-1} \label{eq:consistency}
\end{align}
where Eq. \ref{eq:conservation} imposes people conservation in the neighborhood of a location $j$ (Figure \ref{fig:flows}) across consecutive frame intervals, and Eq. \ref{eq:consistency} enforces the spatio-temporal symmetry of the flows, i.e. the people should move in the opposite direction when the time flows backwards. More details can be found in \cite{liu2020estimating}.

With this weakly-supervised learning framework, the only missing piece is the function that regresses the flows. In particular, this function is a deep neural network  $\mathcal{R}(\boldsymbol{I}^{t-1}, \boldsymbol{I}^{t}, \theta)$ that outputs the flow $\boldsymbol{f}^{t-1,t}$ given two consecutive images in input. Its parameters $\theta$ are optimized during the training process by enforcing the constraints in Eq. \ref{eq:conservation} and \ref{eq:consistency}. In the next paragraph, we propose to use a spatio-temporal attentive network as  $\mathcal{R}$.

\begin{figure*}[t]
  \centering
  \includegraphics[page=3,width=0.93\linewidth]{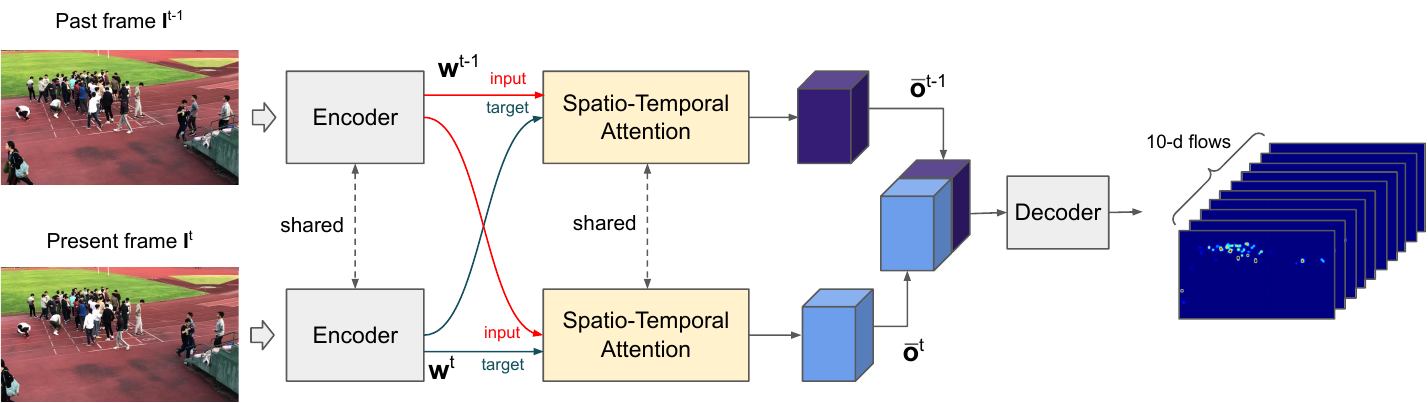}
  \caption{The overall spatio-temporal attentive regressor architecture.}
  \label{fig:overall-architecture}
\end{figure*}

\begin{figure}[t]
  \centering
  \includegraphics[page=4,width=0.95\linewidth]{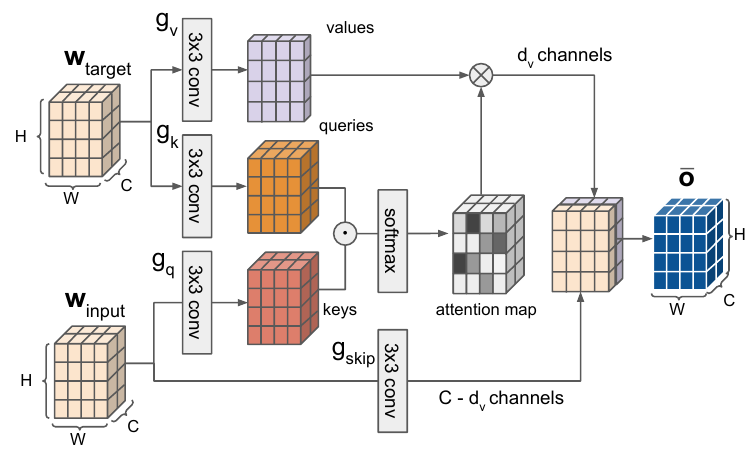}
  \caption{Inner architecture of the spatio-temporal fusion module.}
  \label{fig:attention-architecture}
\end{figure}

\subsection{The Attentive Flow Regressor}

\label{sec:network}
The proposed network resembles a convolutional encoder-decoder architecture, which takes two consecutive RGB images in input and produces the predicted flows. Specifically, the encoder $\mathcal{E}$ processes the input images $\boldsymbol{I}^{t-1}$ and $\boldsymbol{I}^{t}$ to obtain the corresponding internal feature maps $\boldsymbol{w}^{t-1}, \boldsymbol{w}^{t} \in \mathbb{R}^{W\times H\times C}$, where $(W, H, C)$ are the width, the height and the number of channels, respectively. Formally:
\begin{align}
    \boldsymbol{w}^{t-1} &= \mathcal{E}(\boldsymbol{I}^{t-1})\\
    \boldsymbol{w}^{t} &= \mathcal{E}(\boldsymbol{I}^{t})
\end{align}

These feature maps are then composed together using an aggregator function $\tilde{\boldsymbol{w}} = \mathcal{A}(\boldsymbol{w}^{t-1}, \boldsymbol{w}^{t}$), which outputs a feature map $\tilde{\boldsymbol{w}} \in \mathbb{R}^{W\times H\times C'}$, with the same spatial resolution but possibly a different number $C'$ of channels.
In the end, the final flows are obtained by applying the decoder to the aggregated feature maps: $\boldsymbol{f}^{t} = \mathcal{D}(\tilde{\boldsymbol{w}}) \in \mathbb{R}^{W\times H\times 10}$. Notice that the output flow is 10-dimensional, as there are ten possible directions in which a person can move inside the frame (nine locations in the neighborhood including the starting cell, as depicted in Figure \ref{fig:flows}, plus one cell representing \textit{the rest of the world} at the edges of the image).

In the original formulation in \cite{liu2020estimating}, the aggregator function is a straightforward concatenation along the channels dimension: $\mathcal{A} = [\cdot, \cdot]^C$, which therefore outputs $\tilde{\boldsymbol{w}} \in \mathbb{R}^{W\times H\times 2C}$.
In this work, 
we propose an attentive spatio-temporal aggregation module that can produce more space-time aware feature maps $\tilde{\boldsymbol{w}}$, which, in turn, provide better flows. 

The idea is to employ the feature maps coming from one of the two input frames to condition the features of the other frame. We denote the feature maps from the two frames as $\boldsymbol{w}_{\text{input}}$ and $\boldsymbol{w}_{\text{target}}$, where \textit{input} refers to the visual information that conditions the \textit{target} one in the attention mechanism.
In particular, we used an approach similar to the convolutional self-attention by \cite{bello2019attention}.
Specifically, we initially derive other three feature maps from the two input maps produced by the encoder, namely $\boldsymbol{v}=g_v(\boldsymbol{w}_{\text{target}}) \in \mathcal{R}^{W\times H \times d_v}$ and $\boldsymbol{q}=g_q(\boldsymbol{w}_{\text{target}}) \in \mathcal{R}^{W\times H \times d_k}$ from $\boldsymbol{w}_{\text{target}}$, and $\boldsymbol{k}=g_k(\boldsymbol{w}_{\text{input}}) \in \mathcal{R}^{W\times H \times d_k}$ from $\boldsymbol{w}_{\text{input}}$. These $\boldsymbol{v}, \boldsymbol{k}, \boldsymbol{q}$ are the \textit{values}, the \textit{keys}, and the \textit{queries} respectively, used to drive the Transformer-like attention mechanism. They are produced from the input feature maps using three different convolutional layers $g_v, g_k, g_k$ having a $3\times 3$ kernel and padding=1 to leave the spatial resolution untouched.

The output, as in the Transformer \cite{vaswani2017attention} attention mechanism, is computed through the scaled dot-product attention:
\begin{align}
    \boldsymbol{o} &= \text{softmax}(\frac{\boldsymbol{q} \boldsymbol{k}^{\top}}{\sqrt{d_k}})\boldsymbol{v} \\
    &= \text{softmax}(\frac{g_q(\boldsymbol{w}_{\text{target}}) g_k(\boldsymbol{w}_{\text{input}})^{\top}}{\sqrt{d_k}})g_v(\boldsymbol{w}_{\text{target}})
\end{align}

Note that, differently from the original Transformer-like attention, we empirically found that computing values from the target sequence --- instead of the input sequence --- led to better results. 
Still, the core idea aims at incorporating the context from one frame into the feature map of the other frame.

To account for the original feature maps after the attentive processing, we concatenate a transformation of the original feature maps $\boldsymbol{w_\text{input}}$ to the output $\boldsymbol{o}$ along the channel dimension:
\begin{align}
    \boldsymbol{\bar{o}} &= [g_{\text{skip}}(\boldsymbol{w_\text{input}}), \boldsymbol{o}]^C
\end{align}
where $g_{\text{skip}}(\cdot)$ is a convolution operation that does not change the spatial resolution, and outputs $C-d_v$ channels so that $\boldsymbol{\bar{o}}$ has again C channels.
We denote the above-described spatio-temporal attentive fusion module as $\boldsymbol{\bar{o}} = \mathcal{ST}(\boldsymbol{w}_{\text{input}}, \boldsymbol{w}_{\text{target}})$ (see Figure \ref{fig:attention-architecture}). Now, the final goal is to condition the past frame $\boldsymbol{I}^{t-1}$ given the present frame $\boldsymbol{I}^{t}$, and vice-versa. For this reason, we compute the two symmetric attentional feature maps, by simply swapping the two frames in input:
\begin{align}
    \boldsymbol{\bar{o}}^{t} &= \mathcal{ST}(\boldsymbol{w}^{t-1}, \boldsymbol{w}^{t})\\
    \boldsymbol{\bar{o}}^{t-1} &= \mathcal{ST}(\boldsymbol{w}^{t}, \boldsymbol{w}^{t-1})
\end{align}
At this point, using plain concatenation along the channel direction, we merge the obtained attentive feature maps obtaining the final spatio-temporal aware feature map: $\boldsymbol{\bar{w}} = [\boldsymbol{\bar{o}}^{t-1}, \boldsymbol{\bar{o}}^{t}]^C$.
Finally, as in the original encoder-decoder convolutional architecture, the 10-dimensional output flow $\boldsymbol{f}^t$ is obtained by applying the decoder to this last feature map: $\boldsymbol{f}^{t} = \mathcal{D}(\tilde{\boldsymbol{w}})$. The overall architecture is shown in Figure \ref{fig:overall-architecture}.

\section{Experimental Evaluation}
\label{sec:experiments}
This section describes the experiments performed to validate our approach and discusses the obtained results. As benchmark, we exploit the widely-used FDST dataset \cite{DBLP:conf/icmcs/FangZCGH19}. It consists of 100 videos gathered from 13 different scenarios. A total of 150,000 frames have been extracted, thus representing one of the largest and most diverse collections of real-world images suitable for this task. 
Annotations are expressed using dots localizing peoples' heads, as usual for the counting task, for a total of 394,081 labeled pedestrians. We follow the same setting as in \cite{DBLP:conf/icmcs/FangZCGH19}, considering 60 videos (9,000 frames) as the training set and the remaining 40 videos (6,000 frames) as the test set.

In the first stage of the evaluation, we compare the counting errors obtained using our proposed solution against the framework introduced in \cite{liu2020estimating, 9508171}, and other recent state-of-the-art counting solutions present in the literature. On the other hand, in the second part of our experiments, we also assess the ability to correctly localize the counted pedestrians.

\subsection{Comparison with the State-of-the-art}
Here, we compare our solution in terms of counting with other state-of-the-art methodologies. Following standard counting benchmarks, we used the Mean Absolute Error (MAE) and the Root Mean Square Error (RMSE) to measure the counting performance. Specifically, they are defined as:

\begin{equation} \label{eq:mae}
\text{MAE} = \frac{1}{N} \sum_{n=1}^{N} \left| c_\text{gt}^{n} - c_\text{pred}^{n} \right|\,,
\end{equation}

\begin{equation}
\label{rmse_def}
RMSE = \sqrt{\frac{1}{N} \sum_{n=1}^{N} (c_{gt}^{n} - c_{pred}^{n})^{2}}.
\end{equation}

\noindent where $N$ is the number of test images, $c_\text{gt}^{n}$ is the actual count (i.e., the ground truth), and $c_\text{pred}^{n}$ is the predicted count of the $n$-th image. It is worth noting that, as a result of the squaring of each difference, the RMSE effectively penalizes large errors more heavily than small ones, and so it is more useful when outliers are particularly undesirable.

We report our quantitative results obtained on the FDST dataset in Table \ref{tab:comparison_mae_sota}. We called our solution \textit{People Flow Temporal Fusion (TF)}, to distinguish it from the original framework \textit{People Flow Plain Concatenation (PC)}. We divided the training set into train and validation splits, considering the $80\%$ and the $20\%$ of the available data, respectively. To mitigate the overfitting problem, we considered training and validation images belonging to different video sequences (12 sequences for validation and the remaining 48 for the training). For better capturing movement between consecutive images, we sampled the previous and the consecutive frames by setting an offset of 5 frames. More, we performed data augmentation randomly, applying common transformations such as horizontal flipping, cropping, and normalization. We repeated the experiments using our solution three times, reporting the mean. Finally, images are resized to a dimension of $640 \times 360$ pixels (width and height, respectively). As can be seen, our approach outperforms the competing methods. In particular, we lowered the MAE by about $5\%$ compared to the baseline framework, with a comparable RMSE value.

\begin{table}[tbp] 
    \caption{\textbf{Comparison with SOTA on FDST dataset.} 
    We obtained SOTA results in terms of MAE and we lowered the error by $5\%$ compared to the original framework.
    }
    \medskip
    \centering
    \newcolumntype{C}{>{\hsize=0.8\hsize\centering\arraybackslash}X}
    \newcolumntype{B}{>{\hsize=1.6\hsize}X}
    \begin{tabularx}{\linewidth}{BCCC}
    \toprule
     Model & Temporal & MAE $\downarrow$ & RMSE $\downarrow$ \\
    \midrule
    ConvLSTM \cite{xiong2017spatiotemporal} & \checkmark & 4.48 & 5.82 \\
    WithoutLST \cite{fang2019locality} & & 3.87 & 5.16 \\ 
    MCNN \cite{multi_column} & & 3.77 & 4.88 \\ 
    LST \cite{fang2019locality} & \checkmark & 3.35 & 4.45 \\
    CAN \cite{can} & & 2.44 & 2.96 \\
    People Flow PC \cite{liu2020estimating, 9508171} & \checkmark & 2.17 & \textbf{2.62} \\
    \midrule
    People Flow TF (Our) & \checkmark & \textbf{2.07} & 2.69 \\
    \bottomrule
    \end{tabularx} \\[1ex]
    \label{tab:comparison_mae_sota}
\end{table}

\begin{figure}[t]
  \centering
  \includegraphics[page=6,width=0.95\linewidth]{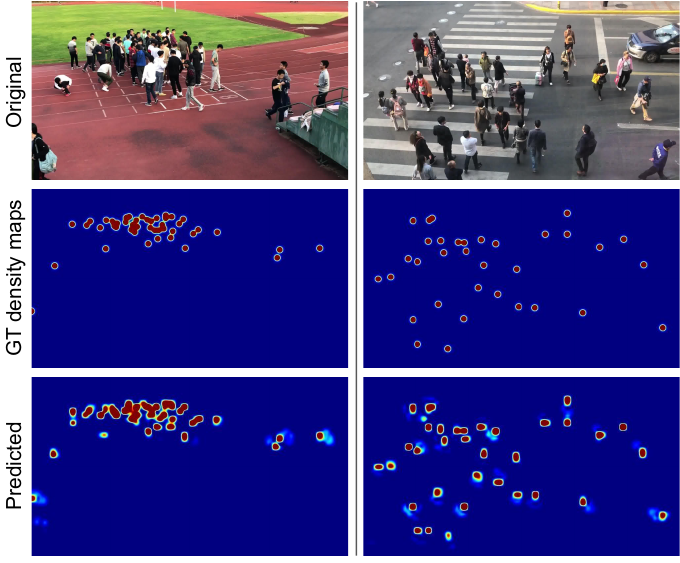}
  \caption{Some qualitative examples showing the predicted density maps.}
  \label{fig:qualitative}
\end{figure}

\subsection{Localization Analysis}
Although the MAE is a fair metric for establishing a comparative in terms of count error, it can often lead to masking erroneous estimations. A potential pitfall of the counting approaches is that the models may miss hard-to-detect instances. To compensate for these missed detections and estimate the correct count, they may falsely mark background sub-regions having similar regional image properties as possible object instances instead. The reason is that the MAE does not take into account \textit{where} the estimations have been done in the images. 
In this section, we conduct experiments to assess the ability of our solution to localize the counted pedestrians correctly, comparing the obtained results against the framework introduced by \cite{liu2020estimating, 9508171}. Specifically, we consider the \textit{Grid Average Mean absolute Error (GAME)} \cite{ExtremelyTrancos}, a hybrid metric that simultaneously considers the object count and the estimated locations of the persons. It is computed by sub-dividing the image in $4^L$ non-overlapping regions and summing the MAE computed in each of these sub-regions:
\begin{equation}
\label{game_def}
GAME(L) = \frac{1}{N} \sum_{n=1}^{N} (\sum_{l=1}^{4^{L}} |c_{gt}^{l} - c_{pred}^{l}|),
\end{equation}
where $N$ is the total number of test images, $c_{pred}^{l}$ is the estimated count in a region $l$ of the n-th image, and $c_{gt}^{l}$ is the ground truth for the same region in the same image. The higher $L$, the more restrictive the GAME metric will be. 

In this work, we considered a grid with a fixed dimension of $80 \times 45$ (width and height, respectively), corresponding to the spatial resolution of the feature maps produced by the encoder $\mathcal{E}$ fed with input images of $640 \times 360$ pixels. Table \ref{tab:localization_analysis} shows our quantitative results obtained on the FDST dataset exploiting our solution \textit{People Flow Temporal Fusion (TF)} compared with the original framework \textit{People Flow Plain Concatenation (PC)}. We repeated the experiments three times, reporting the mean. Specifically, concerning the original framework, we performed three different experiments considering the model publicly provided \footnote{\url{https://github.com/weizheliu/People-Flows}} by the authors of \cite{liu2020estimating, 9508171}, and two models obtained re-training from scratch the original network using different seeds (as illustrated in the Table, the mean concerning the MAE is fully comparable with the one reported in \cite{liu2020estimating, 9508171}). As can be seen, our approach outperforms the original methodology, lowering the GAME by about $7.5\%$, thus suggesting that our solution provides better performance not only in terms of count errors but that it is also capable of better localizing the found pedestrians. In Figure \ref{fig:qualitative} we report some qualitative outputs from our approach.

\begin{table}[tbp] 
    \caption{\textbf{Localization Analysis.} We lowered the \textit{GAME} by about $7.5\%$ compared to the original framework on the FDST dataset.
    }
    \medskip
    \centering
    \newcolumntype{C}{>{\hsize=0.75\hsize\centering\arraybackslash}X}
    \newcolumntype{B}{>{\hsize=1.5\hsize}X}
    \begin{tabularx}{\linewidth}{BCC}
    \toprule
     Model & MAE $\downarrow$ & GAME $\downarrow$ \\
    \midrule
    People Flow PC$^*$ & 2.17 & 16.00 \\    
    \midrule
    People Flow TF (Our) & \textbf{2.07} & \textbf{14.81} \\
    \bottomrule
    \end{tabularx} \\[1ex]
    \raggedright
    \footnotesize
    * Retrained in this work.
    \label{tab:localization_analysis}
\end{table}

\section{Conclusions and Future Work}
\label{sec:conclusion}
This paper proposed an attentive neural network to estimate the number of pedestrians in videos from surveillance cameras. In particular, we extended a promising method that estimates people flows to obtain people densities, which are then integrated to get the final count. To create a more suitable spatial and temporal context for predicting the flows, we proposed a spatio-temporal attentive network, which can contextualize the features maps from the present with those from the past frame vice-versa. In our experiments on the largely-used FDST dataset, we demonstrated the effectiveness of our architecture. We obtained a considerable improvement in counting performance compared to other state-of-the-art approaches in video counting. Furthermore, through the GAME metric, we demonstrated that our method achieves a more precise localization of the pedestrians inside the frame than the network without the spatio-temporal attentive module. 

In the future, we plan to extend the experimentation to other datasets, also employing virtual data to train the network, to avoid manual annotation of large dot-annotated pedestrian videos. Furthermore, we plan to use domain adaptation techniques to fill the well-known \textit{domain-gap} existing between the different monitored scenarios.

\bibliography{biblio.bib}
\bibliographystyle{IEEEtran}

\end{document}